\documentclass[10pt,twocolumn,letterpaper]{article}

\usepackage{iccv}              % To produce the CAMERA-READY version

\usepackage{multirow}

\definecolor{iccvblue}{rgb}{0.21,0.49,0.74}
\usepackage[pagebackref,breaklinks,colorlinks,allcolors=iccvblue]{hyperref}

\def\paperID{3127} % *** Enter the Paper ID here
\def\confName{ICCV}
\def\confYear{2025}

\title{EGVD: Event-Guided Video Diffusion Model for Physically Realistic \\ Large-Motion Frame Interpolation}

\author{
    Ziran Zhang\textsuperscript{1,2} \quad 
    Xiaohui Li\textsuperscript{2,3} \quad 
    Yihao Liu\textsuperscript{2} \quad 
    Yujin Wang\textsuperscript{2} \quad \\
    Yueting Chen\textsuperscript{1} \quad 
    Tianfan Xue\textsuperscript{4,2*} \quad 
    Shi Guo\textsuperscript{2*}\\[0.5em]
    \textsuperscript{1}Zhejiang University \quad
    \textsuperscript{2}Shanghai AI Laboratory \quad \\
    \textsuperscript{3}Shanghai Jiao Tong University \quad
    \textsuperscript{4}The Chinese University of Hong Kong\\[0.3em]
    \textsuperscript{*}Corresponding authors\\[0.5em]
    \url{https://github.com/OpenImagingLab/EGVD}
}

\begin{document}
%\maketitle
\twocolumn[{
\renewcommand\twocolumn[1][]{#1}
\maketitle
\vspace{-30pt}
\begin{center}
    \centering
    \captionsetup{type=figure}
    \includegraphics[width=\linewidth]{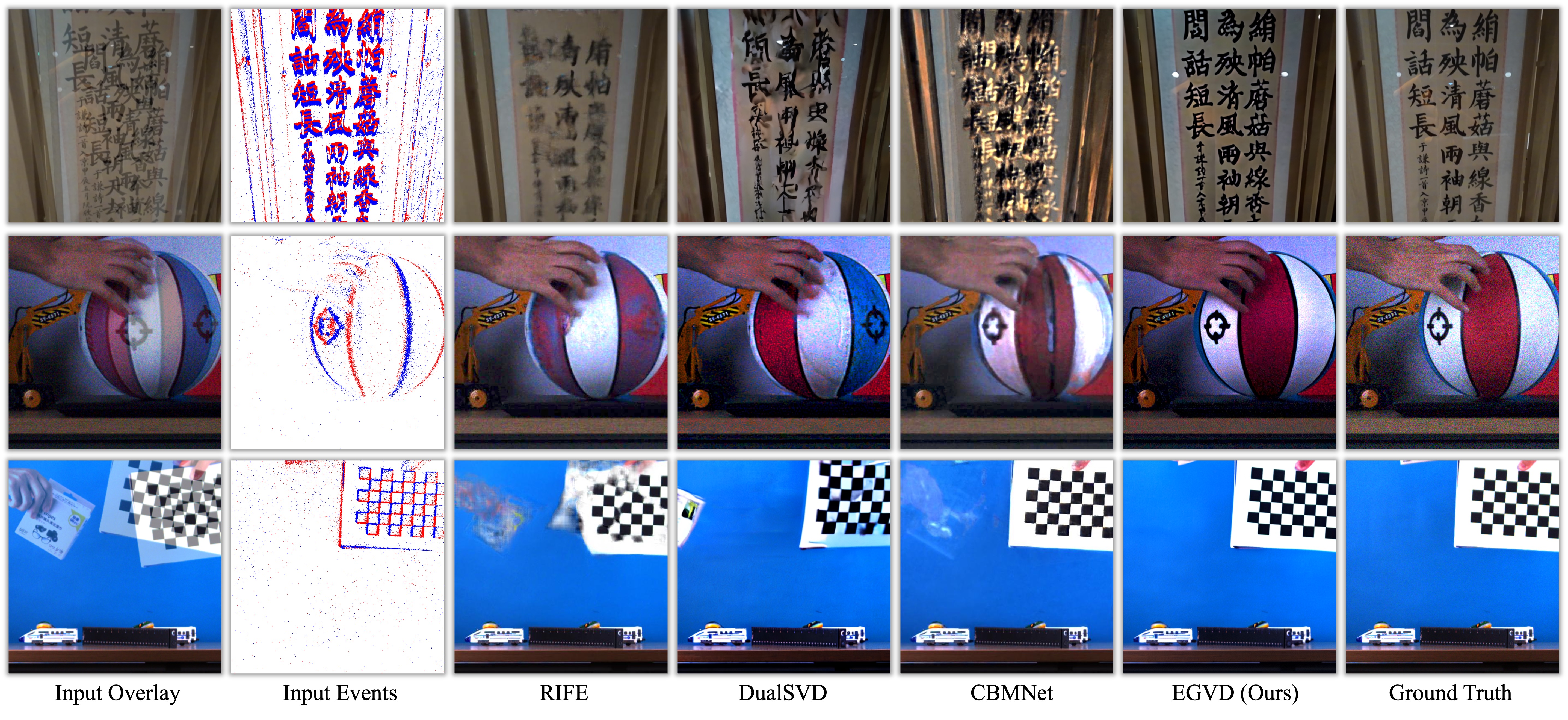}
    \vspace{-20pt}
    \captionof{figure}{Visual comparisons of video frame interpolation (VFI) results across diverse scenes. The top row presents synthetic data generated from a 240fps DJI Action4 video, downsampled to 30fps with simulated event data. The second row shows real-world data captured in low-light conditions, while the third row features real-world data under normal illumination with large motion. We compare RIFE~\cite{huang2022real} (RGB-based VFI), DualSVD~\cite{wang2024generative} (RGB-based VFI with a diffusion model), CBMNet~\cite{kim2023event} (event-based VFI), and our proposed EGVD, which integrates event information within a diffusion-based framework. Our method not only achieves superior interpolation performance, producing sharper reconstructions, but also demonstrates strong generalization and robustness to large motion. \textbf{See the supplementary video for video results.}}
    \label{fig:teaser}
\end{center}%
}]

% \begin{abstract}
% High-speed imaging is essential in computational photography, with applications in dynamic scene capture, turbulence visualization, and fluorescence lifetime imaging. Event cameras, with their high temporal resolution, offer a promising solution for event-based VFI (Event-VFI), yet existing methods face challenges in handling large motion and varying lighting conditions due to limited real-world training data and the ill-posed nature of the problem. To address these challenges, we propose a novel Multi-Modal Motion Condition (MMC) Generator, which enhances motion modeling while maintaining compatibility with the Video Stable Diffusion (SVD) framework. We construct a diverse training dataset by integrating existing Event-VFI datasets and supplementing them with synthetic data, including GoPro sequences and a newly collected 240 FPS dataset. MMC generates conditioning embeddings aligned with the VAE-encoded ground-truth frame, ensuring seamless integration into the RGB domain. Additionally, spatial attention and 3D convolutional layers improve motion representation. We adopt a two-stage training strategy, first training the MMC generator independently, followed by fine-tuning SVD’s attention mechanism to adapt it to Event-VFI while preserving its video priors. Experimental results demonstrate that our approach achieves state-of-the-art interpolation performance, particularly in large-motion and low-light scenarios, offering a stable and efficient solution for Event-VFI.
% \end{abstract}
\vspace{-20pt}
\begin{abstract}
Video frame interpolation (VFI) in scenarios with large motion remains challenging due to motion ambiguity between frames. While event cameras can capture high temporal resolution motion information, existing event-based VFI methods struggle with limited training data and complex motion patterns. In this paper, we introduce Event-Guided Video Diffusion Model (EGVD), a novel framework that leverages the powerful priors of pre-trained stable video diffusion models alongside the precise temporal information from event cameras. Our approach features a Multi-modal Motion Condition Generator (MMCG) that effectively integrates RGB frames and event signals to guide the diffusion process, producing physically realistic intermediate frames. We employ a selective fine-tuning strategy that preserves spatial modeling capabilities while efficiently incorporating event-guided temporal information. We incorporate input-output normalization techniques inspired by recent advances in diffusion modeling to enhance training stability across varying noise levels. To improve generalization, we construct a comprehensive dataset combining both real and simulated event data across diverse scenarios. Extensive experiments on both real and simulated datasets demonstrate that EGVD significantly outperforms existing methods in handling large motion and challenging lighting conditions, achieving substantial improvements in perceptual quality metrics (27.4\% better LPIPS on Prophesee and 24.1\% on BSRGB) while maintaining competitive fidelity measures. Code and datasets available at: https://github.com/OpenImagingLab/EGVD.
 % The code and datasets will be made available once the paper is accepted.
% Our unified model, trained on this diverse data combination, generalizes well across various scenarios without requiring dataset-specific optimization, establishing a new state-of-the-art for event-guided video frame interpolation that bridges the gap between conventional cameras and specialized high-speed imaging systems.
\end{abstract}    
\section{Introduction}
\label{sec:intro}

High-speed imaging plays a crucial role in computational photography, enabling applications such as capturing highly dynamic scenes~\cite{yang2022high}, turbulence visualization~\cite{bai2021predicting}, fast fluorescence lifetime imaging~\cite{lee2019coding}, and high-frequency vibration analysis~\cite{liu2005motion}. However, the widespread use of high-speed cameras is limited by their high cost and the need for intense illumination, which restricts their practicality in many real-world scenarios. In contrast, video frame interpolation (VFI) methods~\cite{bao2019depth, sim2021xvfi, huang2022real, wu2022optimizing} that generate high-frame-rate videos from low-frame-rate recordings offer a more accessible alternative, especially when using conventional RGB cameras. Nevertheless, due to the lack of motion information between frames, traditional RGB-based VFI methods often struggle to handle complex, non-linear motions.

Event cameras, which capture changes in the scene at extremely high temporal resolutions, provide a promising solution to these challenges~\cite{lichtsteiner2008128, brandli2014240, guo20233}. By capturing motion information between frames, event-based VFI (Event-VFI) has demonstrated significant improvements over traditional methods, particularly in scenes with non-linear motion~\cite{tulyakov2021time, tulyakov2022time, kim2023event}. Despite these advances, existing Event-VFI techniques still struggle with large-motion scenarios and varying lighting conditions~\cite{chen2024repurposing}, as shown in \cref{fig:teaser}. These challenges primarily stem from the limited availability of real-world event training data and the ill-posed nature of the interpolation problem in such complex contexts. Therefore, achieving visually pleasing results for Event-VFI under large motion and diverse lighting conditions remains an open challenge.

%Recently, video stable diffusion (SVD) models have gained traction due to their generative capabilities in various tasks, including video colorization~\cite{liu2023video}, RGB-based VFI~\cite{feng2024explorative}, and video super-resolution~\cite{li2025diffvsr}. These models leverage strong priors and have shown promising results in video generation. Inspired by this, \cite{chen2024repurposing} introduced SVD into Event-VFI, showing improvements, particularly in handling large-motion scenes. However, their method relies on separately generating frames for the initial and final inputs and subsequently merging them, which leads to increased inference time and training instability. \textcolor{red}{[lyh: This section mentions the problems with the previous methods in inference and training. Has our model solved the problem of inference efficiency accordingly? If yes, please mention how we solved it. If not, perhaps we can elaborate on the additional issues with the previous method]}

Recently, the development of video generation models has significantly advanced generative models. Video generation models, such as video stable diffusion (SVD) models~\cite{blattmann2023stable}, contain billions of parameters and are trained on millions of high-quality video clips, embedding strong video priors. Leveraging these video priors, diffusion-based pre-trained models have been applied to various low-level vision tasks, including video colorization~\cite{liu2023video}, RGB-based video frame interpolation (VFI)~\cite{feng2024explorative}, and video super-resolution\cite{li2025diffvsr}, significantly improving visual quality. As shown in \cref{fig:teaser}, diffusion-based RGB-VFI methods, such as DualSVD~\cite{wang2024generative}, can generate visually pleasing results in the absence of inter-frame motion information. This demonstrates the potential of diffusion priors in handling large motions in VFI tasks. Still, naively using diffusion model in interpolation may lead visually pleasing but incorrect motion, as shown in the last row of \cref{fig:teaser}.
% Motivated by this, we leverage the prior knowledge of SVD to address the challenges of Event-VFI under large motion and diverse lighting conditions.

Therefore, in this work, we propose an event-guided video diffusion model (EGVD), which both utilize the strong prior from video diffusion model, but also ensure the correctness of reconstruction result.%for physically realistic large-motion frame interpolation. 
The key factor of EGVD is modeling event motion information along with adjacent RGB frames as a conditional signal for SVD finetuning. To achieve effective condition modeling, we design a Multi-Modal Motion Condition Generator (MMCG), which integrates the motion cues from events and RGB frames into the RGB domain. It formulates the condition generator as a coarse Event-VFI process, where its output approximates the VAE encoding of the interpolated frame. The generating conditions that resemble the coarse interpolation results brings them closer to the original input format of SVD, reducing optimization complexity of finetuning SVD. 

% The obtained condition is used as the input to SVD, replacing the original starting frame, and a pre-trained stable video diffusion (SVD) model is fine-tuned for Event-VFI based on this condition. 
With the novel condition calculated from input event stream, the next step is to finetune a pre-trained stable video diffusion (SVD) model is based on this condition. To reduce the optimization complexity of fine-tuning SVD, we propose a two-stage optimization strategy: the condition generator is first trained in a supervised manner as a separate stage, allowing only the SVD model to be fine-tuned in the later stage. Since fine-tuning SVD for a new task requires a large amount of data, we curate a dataset by gathering both real and synthetic Event-VFI scenes from publicly available sources, as well as capturing our own simulated scenes. The dataset contains 122,810 frames across 400 scenes, which we use for model training. Extensive experimental results demonstrate that our approach outperforms existing methods, particularly in large-motion and low-light scenarios.

Our contributions are summarized as follows:
%In this work, 
%To address the challenges faced by SVD in Event-VFI, we propose an Event-Guided Video Diffusion Model for large-motion frame interpolation. Our method introduces the Multi-Modal Motion Condition (MMC) Generator, which effectively incorporates event-based motion information into the SVD framework, ensuring that the temporal dynamics captured by event cameras are used to guide the generation of intermediate frames. By fine-tuning SVD with the event-based conditioning, we achieve efficient and physically realistic frame interpolation in large-motion scenarios. Our contributions are as follows:
\begin{itemize}
    \item We introduce a novel Multi-Modal Motion Condition Generator(MMCG), which integrates event information into the SVD framework to improve the interpolation of large motions.
    \item We propose a two-stage training strategy, which first trains the conditioning generator independently, followed by fine-tuning the SVD model to adapt to Event-VFI.
    \item We construct a diverse and comprehensive training dataset, combines real-world and synthetic event-RGB data, improving the generalization ability of our model.
    \item Extensive experimental results demonstrate that our approach outperforms existing methods, particularly in large-motion and low-light scenarios.
\end{itemize}

\begin{figure*}
    \centering
    \includegraphics[width=\linewidth]{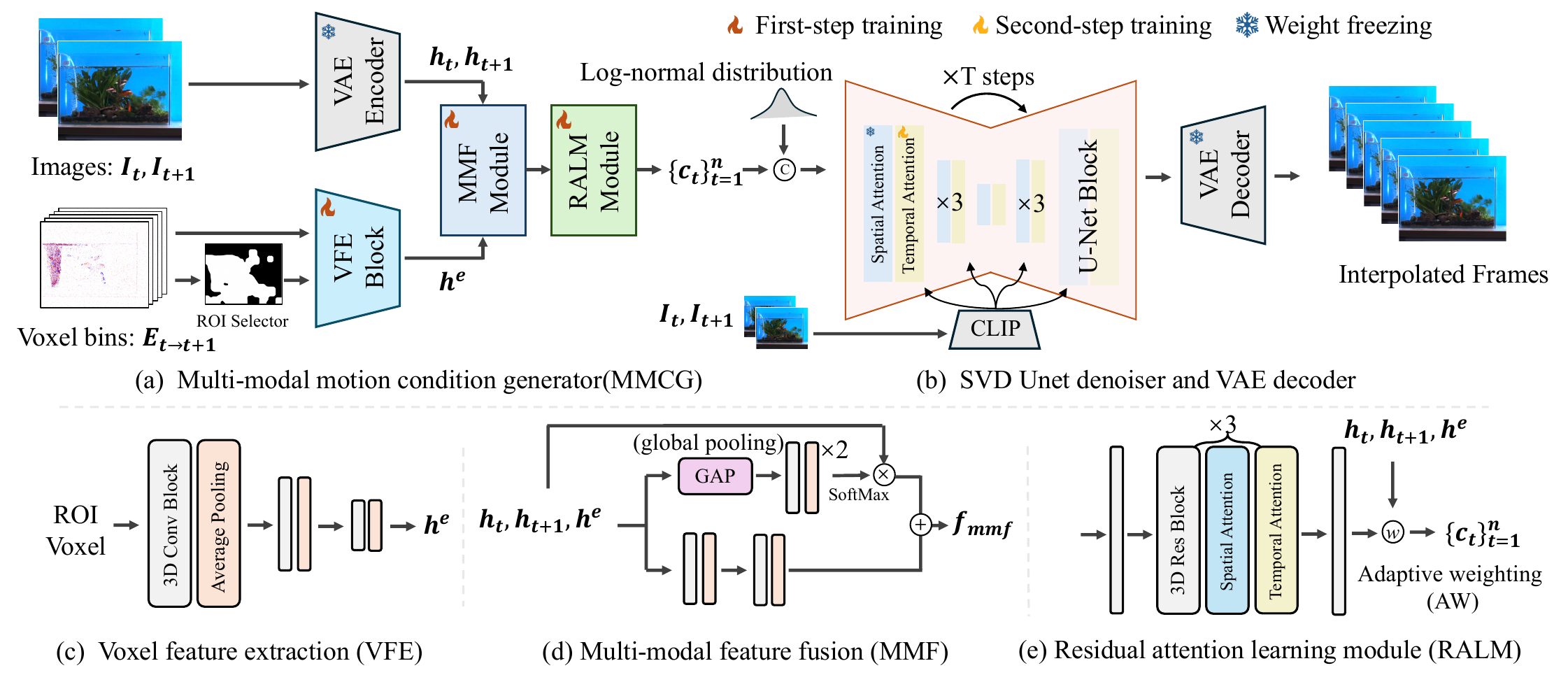}
    \caption{The illustration of our Event-Guided Video Diffusion Model for physically realistic large-motion frame interpolation. }
    \label{fig:framework}
\end{figure*}
\section{Related work}
\subsection{RGB-based video frame interpolation}
Video frame interpolation is a fundamental yet challenging task aimed at generating intermediate frames between consecutive video frames. Some learning-based methods directly generate inter-frames using techniques such as adaptive separable convolution~\cite{niklaus2017video}, phase decomposition~\cite{meyer2018phasenet}, and transformer-based structures~\cite{shi2022video}. Another class of methods~\cite{bao2019depth,sim2021xvfi,huang2022real,wu2022optimizing} relies on the explicit estimation of optical flow during interpolation. However, RGB-VFI methods often experience degraded performance when handling complex and large motion, primarily due to the lack of accurate motion information between consecutive RGB frames. Recently, diffusion-based models~\cite{jain2024video,danier2024ldmvfi,huang2024motion}, which have seen significant success in generative modeling~\cite{ho2020denoising,rombach2022high}, have been applied to RGB-based video frame interpolation, leading to substantial improvements, particularly in scenes with large motion. However, due to the absence of detailed motion information, the generated motion often fails to adhere to physical laws and does not align with real-world motion dynamics.

\subsection{Event-based video frame interpolation}
Event cameras, which capture intensity changes of objects, are known for their high frame rates and wide dynamic range~\cite{posch2010qvga,chen2024event,gehrig2024low}, making them particularly beneficial for video frame interpolation (VFI) tasks. As a result, the use of event cameras for VFI (Event-VFI) has gained traction as a promising solution for interpolating complex, non-linear, and large motions~\cite{tulyakov2021time,tulyakov2022time,sun2023refid,kim2023event}. Notable approaches such as Time Lens~\cite{tulyakov2021time} and its enhanced version, Time Lens++~\cite{tulyakov2022time}, have demonstrated strong performance in handling non-linear motion interpolation. In addition, \citet{sun2023refid} proposed REFID, a framework that jointly performs image deblurring and interpolation. Meanwhile, \citet{kim2023event} introduced CBMNet, which incorporates cross-modal asymmetric bidirectional motion field estimation to further improve interpolation accuracy. TimeLens-XL~\cite{ma2024timelens} aims to address large-motion interpolation and is capable of real-time processing. However, due to the ill-posed nature of the interpolation problem in scenes with large motion, previous methods struggle to produce visually pleasing results. To address this issue, we incorporate video diffusion priors to enhance interpolation quality.

\section{Video diffusion based Event-VFI}
To obtain the interpolated frames $\boldsymbol{I_{t+\Delta t}}$ between two neighbor frames $\boldsymbol{I_t}$, $\boldsymbol{I_{t+1}}$ and corresponding events $\boldsymbol{E}$, where $\boldsymbol{\Delta t \in (t, t+1)}$, a video diffusion based Event-VFI framework is designed, as illustrated in \cref{fig:framework}. The key factor of this framework is incorporating the motion information from events and the RGB frames as motion conditions within the stable video diffusion (SVD) framework. Therefore, we first review the background of SVD in \cref{sec:pre-svd}, followed by description on how RGB and event signals control the generation process of SVD in \cref{sec:evs-condition}, along with details on the SVD fine-tuning procedure in \cref{sec:svd-train}. 

\subsection{Stable video diffusion}
\label{sec:pre-svd}
Diffusion models~\cite{rombach2022high} are generative models that add noise to data $\boldsymbol{x_0} \sim p_{\text{data}}(\boldsymbol{x})$ to obtain Gaussian noise $\boldsymbol{x_T} \sim \mathcal{N}(0, \boldsymbol{I})$ and learn a reverse process to progressively remove the noise. $\boldsymbol{I}$ is the identity matrix. For video generation, image-conditioned stable video diffusion (SVD)~\cite{blattmann2023stable} is proposed, which incorporates 3D convolution and temporal attention layers to model temporal dynamics.
% A video $I \in \mathbb{R}^{N \times T \times H \times W}$ is encoded into a latent space $z = \mathcal{E}(I)$, where forward and reverse processes are applied, and the final video is decoded from this latent space. The objective function is:
% %
% \begin{equation}
% L_{\text{noise}} = \mathbb{E}_{\mathcal{E}(x), \epsilon \sim \mathcal{N}(0, 1), t} \left\| \epsilon - \epsilon_\theta (z^t, t, c) \right\|_2^2 ,
% \label{eq:svd-loss}
% \end{equation}
% %
% where $t \in \{0, \dots, T\}$, $\epsilon_\theta$ is a diffusion U-Net with temporal attention, $c$ is the condition, which uses the frame $I_0$ in the default SVD model. 
Since the original SVD generates the video from the first frame, despite having prior knowledge, the motion is not physically consistent. Therefore, we aim to leverage both the high temporal resolution of event cameras and the powerful generative capabilities of diffusion models to address physically realistic large-motion frame interpolation. The core step is how to model the motion information from $\boldsymbol{I_t}$ and $\boldsymbol{I_{t+1}}$ with event voxel signals $\boldsymbol{E}$ as condition $c$ to control the output of SVD.

For model motion condition in RGB-VFI~\cite{feng2024explorative, wang2024generative} and Event-VFI~\cite{chen2024repurposing}, a common approach is to perform image-to-video tasks conditioned on the initial and final frames separately, then merge the intermediate features to create the final interpolated frame. However, this approach requires running two Diffusion U-Nets, which doubles the inference time. Additionally, training two diffusion models introduces complexity and instability to the training process. To address these limitations and efficiently control the SVD for interpolation, we propose the Multi-Modal Motion Condition Generator (MMCG) to explicitly generate motion conditions.

\begin{table*}[t]
\centering
\caption{Quantitative comparison of video interpolation methods across different datasets. 
The best results are highlighted in \textbf{bold}, while the second-best results are underlined. All datasets consist of RGB images with a skip of $\times 3$ to evaluate large motion.}
\label{tab:results}
\begin{tabular*}{\textwidth}{@{\extracolsep{\fill}}llcccccc@{}}
\toprule
Dataset & Method & PSNR$\uparrow$ & SSIM$\uparrow$ & LPIPS$\downarrow$ & MANIQA$\uparrow$ & MUSIQ$\uparrow$ & LIQE$\uparrow$ \\
\midrule
\multirow{5}{*}{Prophesee}  
& SuperSloMo~\cite{jiang2018super} & 21.13 & 0.6569 & 0.2051 & \textbf{0.2278} & \underline{45.03} & \underline{1.451} \\
& RIFE~\cite{huang2022real} & 21.64 & 0.6728 & 0.2276 & 0.1988 & 42.20 & 1.180 \\
& CBMNet~\cite{kim2023event} & \textbf{24.01} & \underline{0.7399} & 0.2512 & 0.2134 & 43.14 & 1.233 \\
& DualSVD~\cite{wang2024generative} & 19.98 & 0.7568 & \underline{0.1958} & 0.2033 & 41.75 & 1.228 \\
& EGVD (Ours) & \underline{22.87} & \textbf{0.7768} & \textbf{0.1422} & \underline{0.2225} & \textbf{46.04} & \textbf{1.475} \\
\midrule
\multirow{5}{*}{BS-RGB}  
& SuperSloMo~\cite{jiang2018super} & 18.22 & 0.5552 & \underline{0.2283} & \underline{0.2830} & \textbf{57.38} & \underline{2.347} \\
& RIFE~\cite{huang2022real} & 18.94 & 0.5859 & 0.2601 & 0.2697 & 54.81 & 1.964 \\
& CBMNet~\cite{kim2023event} & \textbf{21.62} & \textbf{0.6487} & 0.2216 & 0.2595 & 52.84 & 1.956 \\
& DualSVD~\cite{wang2024generative} & 18.67 & \underline{0.6418} & 0.2596 & 0.2322 & 51.52 & 1.920 \\
& EGVD (Ours) & \underline{19.81} & 0.5913 & \textbf{0.1732} & \textbf{0.2831} & \underline{57.31} & \textbf{2.557} \\
\midrule
\multirow{5}{*}{DJI 30fps}  
& SuperSloMo~\cite{jiang2018super} & 17.94 & 0.6082 & 0.3228 & \underline{0.2972} & \underline{61.07} & \underline{2.456} \\
& RIFE~\cite{huang2022real} & 17.88 & 0.6254 & 0.3789 & 0.2675 & 52.73 & 1.830 \\
& CBMNet~\cite{kim2023event} & 16.24 & 0.5486 & 0.4298 & 0.1660 & 43.37 & 1.244 \\
& DualSVD~\cite{wang2024generative} & \underline{18.73} & \underline{0.6457} & \underline{0.2906} & 0.2506 & 54.86 & 2.466 \\
& EGVD (Ours) & \textbf{22.46} & \textbf{0.7555} & \textbf{0.1993} & \textbf{0.3321} & \textbf{61.38} & \textbf{3.096} \\
\midrule
\multirow{5}{*}{GOPRO 240fps}  
& SuperSloMo~\cite{jiang2018super} & 21.32 & 0.6745 & \textbf{0.1309} & 0.2020 & \underline{47.24} & 1.532 \\
& RIFE~\cite{huang2022real} & \underline{21.90} & \underline{0.6910} & \underline{0.1397} & 0.2069 & \textbf{47.55} & 1.523 \\
& CBMNet~\cite{kim2023event} & 13.70 & 0.4103 & 0.4057 & 0.1567 & 30.85 & 1.068 \\
& DualSVD~\cite{wang2024generative} & 18.24 & 0.5655 & 0.2161 & \underline{0.2199} & 45.22 & \underline{1.645} \\
& EGVD (Ours) & \textbf{23.84} & \textbf{0.8039} & 0.1579 & \textbf{0.2207} & 46.35 & \textbf{1.725} \\
\bottomrule
\end{tabular*}
\end{table*}

\subsection{Multi-modal motion condition generator}
\label{sec:evs-condition}
As illustrated in \cref{fig:framework} (a), the MMCG generates motion conditions $\{\boldsymbol{c}_t\}^n$, where $n$ is the number of interpolated frames, from $\boldsymbol{I}_t$, $\boldsymbol{I}_{t+1}$, and the corresponding event signals $\boldsymbol{E}_{t \rightarrow t+1}$. These conditions are then used to control the SVD process. We denote the MMCG as $\mathcal{G}$, with the output motion condition features expressed as:
\begin{equation}
\boldsymbol{c}_t = \mathcal{G}(\boldsymbol{E}_{t \rightarrow t+1}, \boldsymbol{I}_t, \boldsymbol{I}_{t+1}).
\end{equation}
By incorporating MMCG as the motion condition controller within the SVD framework, we develop our Event-Guided Video Diffusion Model (EGVD), as illustrated in Fig.~\ref{fig:framework}. The MMCG consists of three key components: feature extraction, multi-modal feature fusion, and condition generation, as described below.

\textbf{Feature extraction module.}
Feature extraction module is used to encode RGB and event signals into feature space. 
\begin{equation}
\boldsymbol{h_t} = \mathcal{E}_{\text{VAE}}(\boldsymbol{I_t}), \quad \boldsymbol{h^e} = \mathcal{E}_{\text{VFE}}(\boldsymbol{E}, \mathrm{M}(\boldsymbol{E})),
\end{equation}
where $\mathcal{E}_{\text{VAE}}(\cdot)$ is the pre-trained VAE encoder which is frozen in the training process, and $\mathcal{E}_{\text{VFE}}$ is the voxel feature extraction module, which consists of 3D convolutions with LeakyReLU activations to model temporal information. 

We employ a voxel grid representation with $8$ temporal bins to encode fine-grained event features $\boldsymbol{E}$ for each frame\cite{zhu2024continuous}. We implement an ROI selection mechanism to prioritize computation on motion-relevant regions. The event stream is first normalized ($\boldsymbol{E'} = |\boldsymbol{E}| \, / \max(|\boldsymbol{E}|)$) and smoothed with a Gaussian filter ($\boldsymbol{E''} = \boldsymbol{G_\sigma} \ast \boldsymbol{E'}$). Binary masks are generated by thresholding ($\boldsymbol{B}=1$ if $\boldsymbol{E''} > 0.01$ else 0), followed by morphological dilation and median filtering to create coherent motion regions ($\mathrm{M}(\boldsymbol{E}) = \boldsymbol{M_{\text{kmed}}}(\boldsymbol{B} \oplus \boldsymbol{K}_{\text{dilate}})$). The final mask integrates ROIs across all temporal channels, concentrating the model's attention on areas with significant event activity while reducing computational overhead in static regions.

\textbf{Multi-modal feature fusion.}
Multi-modal feature fusion (MMF) aims to fuse information from both RGB and event modalities. The MMF process is defined as:
%
% \begin{equation}
%     f_{\text{fuse}}, w = \text{MMF}(h_t, h_{t+1}, h^e),
% \end{equation}

\begin{equation}
\boldsymbol{f}_{\text{fuse}}, \boldsymbol{w} = \mathrm{MMF}\left(\boldsymbol{h}_t, \boldsymbol{h}_{t+1}, \boldsymbol{h}^e\right),
\end{equation}
where MMF utilizes 3D convolutions to generate the fusion feature $f_{\text{fuse}}$ and the predicted weight vector $\boldsymbol{w} = [\boldsymbol{w}_1, \boldsymbol{w}_2, \boldsymbol{w}_3]$. The final fused feature is computed as:
\begin{equation}
\boldsymbol{f}_{\text{mmf}} = \boldsymbol{w}_1 \boldsymbol{h}_t + \boldsymbol{w}_2 \boldsymbol{h}_{t+1} + \boldsymbol{w}_3 \boldsymbol{h}^e + \boldsymbol{f}_{\text{fuse}}.
\end{equation}
%A detailed description of the MMF network structure is provided in the supplementary material.

% \textbf{Residual attention learning module}
% The residual attention learning module (RALM) generates motion control information $ \{c_t\}_{t=1}^n $ as:
% %
% \begin{equation}
%     \boldsymbol{f}_\text{evs} =\text{RALM}(\boldsymbol{f}_{\text{mmf}}),
% \end{equation}
% %
% where $\text{RALM}(\cdot)$ consists 5个 3D residual blocks, 其中中间三个还包括temporal attention~\cite{rosin2022temporal} and spatial attention~\cite{zhu2019empirical} to increase the modeling ability. 

\textbf{Residual attention learning module.}
The Residual Attention Learning Module (RALM) is designed to generate supplementary features for condition control information $\boldsymbol{c}_t$ used in frame interpolation. Specifically, RALM processes the fused feature $ \boldsymbol{f}_{\text{mmf}} $ as input and produces the motion condition feature $ \boldsymbol{f}_{\text{evs}} $ as follows:
\begin{equation}
    \boldsymbol{f}_{\text{evs}} = \text{RALM}(\boldsymbol{f}_{\text{mmf}}).
\end{equation}
As illustrated in Fig.~\ref{fig:framework}(e), the RALM architecture comprises five 3D residual blocks, with the middle three blocks incorporating both temporal attention~\cite{rosin2022temporal} and spatial attention~\cite{zhu2019empirical} mechanisms. These attention mechanisms significantly enhance the module's ability to capture complex motion patterns and spatial dependencies across frames. The output feature $ \boldsymbol{f}_{\text{evs}} $ is subsequently passed to the Adaptive Weighting (AW) module, which utilizes these features to generate the optimal conditioning signals $ \boldsymbol{c}_t $ for the diffusion model.

\textbf{Adaptive weighting.}
For feature fusion across the temporal dimension, we employ a temporal-based dynamic weight allocation. Given a sequence of length $T+1$ (with $T-1$ intermediate frames between two key frames), the weights for frame features at time step $k \in [0,T]$ are:
\begin{equation}
\boldsymbol{w}_{h_t}(k) = k/T, \quad \boldsymbol{w}_{h_{t+1}}(k) =(T-k)/T,
\end{equation}
Please note that the indices $[1, T-1]$ represent the intermediate frames to be interpolated, while $0$ and $T$ correspond to the input frames that serve as the boundary conditions for the interpolation process. Event features $\boldsymbol{f}_{\text{evs}}$ contribute only to intermediate frames:
\begin{equation}
\boldsymbol{w}_{\text{evs}}(k) =
\begin{cases}
    1, & \text{if } k \in \{1,2,...,T-1\} \\
    0, & \text{if } k \in \{0,T\}
\end{cases},
\end{equation}
The final feature for each time step is then obtained through:
% \begin{equation}
% c_t = w_{\text{evs}} \cdot f_{\text{evs}} + w_{h_t} \cdot h_t + w_{h_{t+1}} \cdot h_{t+1}
% \end{equation}
\begin{equation}
\boldsymbol{c}_t = \boldsymbol{w}_{\text{evs}} \cdot \boldsymbol{f}_{\text{evs}} + \boldsymbol{w}_{h_t} \cdot \boldsymbol{h}_t + \boldsymbol{w}_{h_{t+1}} \cdot \boldsymbol{h}_{t+1},
\end{equation}
This weighting strategy not only ensures smooth transitions between frames, but also leverages event information for residual learning to compensate for the incomplete information in the intermediate frames. It is worth noting that for the input frames at $k=0$ and $k=T$, the original information remains unchanged, preserving the integrity of the initial inputs throughout the interpolation process.

% Note that the motion condition $ c_t $ has physical meaning. We aim for $ c_t = \text{VAE-Encoder}(\hat{I}_t) $, where $ \hat{I}_t $ is the interpolated ground-truth. From $ c_t $, we obtain the intermediate frame result $ \text{VAE-Decoder}(c_t) $. As shown in Figure X, due to the absence of prior information from Diffusion, the result for larger motion is not visually pleasing. This highlights the necessity of using Diffusion for Event-VFI in scenarios with significant motion.

\subsection{Training of EGVD  }
\label{sec:svd-train}

\paragraph{Training of MMCG}
Let $\mathcal{I} = \{\boldsymbol{I}_0, \boldsymbol{I}_1, ..., \boldsymbol{I}_N, \boldsymbol{I}_{N+1}\}$ represent the complete video frame sequence, where $\boldsymbol{I}_0$ and $\boldsymbol{I}_{N+1}$ are the input keyframes, and $\{\boldsymbol{I}_1, \boldsymbol{I}_2, ..., \boldsymbol{I}_N\}$ are the intermediate frames to be predicted. Given the input frames $\boldsymbol{I}_0$, $\boldsymbol{I}_{N+1}$, and the event stream $\boldsymbol{E}$ between them, the training objective of the condition generator $\mathcal{G}_{\Theta}$ is to produce feature representations that are as consistent as possible with the complete sequence $\mathcal{I}$ in the latent space. Notably, our model not only predicts the intermediate frames but also preserves the input frames in the output, thus generating latent representations for all $N+2$ frames. The training loss function is defined as the mean squared error in the latent space:
\begin{equation}
\mathcal{L}_{\text{MMCG}}(\boldsymbol{\Theta}) = \frac{1}{N+2} \sum_{i=0}^{N+1} \| \mathcal{E}_{\text{VAE}}(\boldsymbol{I}_i) - \mathcal{G}_{\boldsymbol{\Theta}}(\boldsymbol{E}, \boldsymbol{I}_0, \boldsymbol{I}_{N+1})_i \|_2^2
\label{eq:cond_loss}
\end{equation}
where $\mathcal{E}_{\text{VAE}}$ represents the VAE encoder with fixed parameters, $\boldsymbol{I}_i$ is the $i$-th frame in the sequence, $\mathcal{G}_{\Theta}$ denotes the condition generator with parameters $\Theta$, and $\mathcal{G}_{\Theta}(\boldsymbol{E}, \boldsymbol{I}_0, \boldsymbol{I}_{N+1})_i$ represents the predicted feature for the $i$-th frame by the condition generator.

\paragraph{Denoising model fine-tuning}
\label{subsec:denoising_tuning}

We fine-tune the pre-trained SVD model by focusing only on temporal self-attention modules, specifically the value ($W_v$) and output ($W_o$) projection layers. Our fine-tuning objective function borrows from Elucidated Diffusion Model (EDM)\cite{karras2022elucidating} to ensure training stability by normalizing inputs, outputs, and training targets to maintain consistent standard deviations throughout the training process.

The process begins by adding Gaussian noise $\boldsymbol{N} \sim \mathcal{N}(0, \boldsymbol{I})$ to latent features $\boldsymbol{z}$ with noise level $\sigma_t$ sampled from a log-normal distribution:
\begin{equation}
\tilde{\boldsymbol{z}} = \boldsymbol{z} + \sigma_t  \cdot\boldsymbol{N}, \quad \sigma_t = \exp(\epsilon_t), \quad \epsilon_t \sim \mathcal{N}(0.7, 1.6).
\end{equation}
We normalize inputs to ensure stable network processing:
\begin{equation}
\tilde{\boldsymbol{z}}_{\text{norm}} = \tilde{\boldsymbol{z}}/(\sqrt{\sigma_t^2 + 1}).
\end{equation}
The UNet then predicts the denoised features: $\boldsymbol{z}_{\text{pred}} = \text{UNet}(\tilde{\boldsymbol{z}}_{\text{norm}}, \boldsymbol{c}_t)$.
%\begin{equation}
%\mathbf{z}_{\text{pred}} = \text{UNet}(\tilde{\mathbf{z}}_{\text{norm}}, \mathbf{c}).
%\end{equation}
The denoised representation is calculated using normalized coefficients:
\begin{equation}
\boldsymbol{z}_{\text{denoised}} = \boldsymbol{z}_{\text{pred}} \cdot \frac{-\sigma_t}{\sqrt{\sigma_t^2 + 1}} + \tilde{\boldsymbol{z}} \cdot \frac{1}{\sigma_t^2 + 1}.
\end{equation}
Our noise-level-weighted loss function balances contributions across different noise levels:
\begin{equation}
\mathcal{L}_{\text{denoise}} = \mathbb{E}_{\boldsymbol{z}, \boldsymbol{N}} \left[ \frac{1 + \sigma_t^2}{\sigma_t^2} \cdot \left\| \boldsymbol{z}_{\text{denoised}} - \boldsymbol{z} \right\|_2^2 \right].
\end{equation}
This selective fine-tuning approach, combined with data normalization techniques that stabilize diffusion training, preserves spatial modeling capabilities while efficiently incorporating event-guided temporal information, resulting in significantly reduced training data requirements and accelerated convergence.

\section{Experiment}

\subsection{Dataset}

We evaluate our method on four diverse datasets: Prophesee, BS-ERGB\cite{tulyakov2022time}, DJI 30fps, and GOPRO 240fps\cite{Nah_2017_CVPR}. The Prophesee dataset is a combination of three sub-datasets captured by a Prophesee event camera in tandem with a synchronized RGB camera system. It comprises HQEVFI\cite{ma2024timelens} under normal lighting, EVF-LL\cite{zhang2024sim} in low-light, and ERDS\cite{cho2024tta} under both lighting conditions. For simplicity, we refer to this combined collection as Prophesee. 
% We opted to exclude ERF-X170FPS\cite{kim2023event} from our training and evaluation protocols due to concerns regarding potential temporal alignment inconsistencies that could affect the reliability of our results.

%The DJI 30fps dataset, which we captured using a DJI Action4 camera at 240fps in challenging scenarios, serves as our source data. Using the event simulation method reported by Zhang et al.~\cite{zhang2024sim}, we simulate events in the reverse ISP space with the v2e simulator, ensuring that RIFE is not used during the simulation of large-motion events. This results in an eightfold reduction in the effective frame rate relative to the original footage. In contrast, the GOPRO240fps dataset undergoes a similar simulation process; however, RIFE is applied for eightfold frame interpolation, yielding minimal inter-frame motion.

The GOPRO 240fps dataset primarily consists of global motion, which limits its ability to represent diverse real-world scenes. To address this, we captured the DJI 30fps dataset using a Action4 camera at 240fps in challenging scenes. We simulate events in the reverse ISP space with the v2e simulator, following the method in~\cite{zhang2024sim}, while ensuring that RIFE is not used for large-motion events. This results in an effective frame rate reduced by a factor of eight. In contrast, the GOPRO 240fps dataset undergoes a similar simulation, but RIFE is applied for eightfold frame interpolation, minimizing inter-frame motion.

For training, we use both simulated and real-world data, totaling 400 video sequences: 180 from DJI, 22 from GOPRO, 63 from BS-ERGB, and 135 from Prophesee. The test set includes 11 DJI, 11 GOPRO, 26 BS-ERGB, and 15 Prophesee videos.
%For training, we integrate both simulated and real-world data, assembling a total of 400 video sequences (180 from DJI Simulated Dataset, 22 from GOPRO Simulated Dataset, 63 from BS-ERGB, and 135 from Prophesee). The test set comprises 11 DJI Simulated Dataset, 11 GOPRO Simulated Dataset, 26 BS-ERGB, and 15 Prophesee videos. 
To rigorously evaluate performance under large-motion conditions, we adopt a ``skip-three-frames'' strategy, where every third frame is selected from each video, significantly increasing the frame displacement. Notably, our mixed training strategy combines skip-one and skip-two frame sequences, with skip-three frames randomly incorporated during training.

\subsection{Implementation details}
The proposed Event-Guided Video Diffusion (EGVD) model is implemented in PyTorch and trained on a server equipped with four NVIDIA A100 GPUs. Training proceeds in two stages. In the initial phase, a conditional generator is trained for 30,000 iterations using the Adam optimizer with a fixed learning rate of $1\times10^{-4}$ and a batch size of 4. This is followed by a fine-tuning stage focusing on the temporal self-attention layer within the SVD Unet for an additional 30,000 iterations. The patch size during the training process is 512×512. For inference, we adopt the DDIM sampler with 50 steps to balance generation quality and efficiency, and classifier-free guidance\cite{ho2022classifier} is applied to further refine the output.

% \subsection{Training Data}

%\subsection{Comparison with state-of-the-art methods}

\subsection{Comparison with state-of-the-art methods}

We conduct extensive comparisons against leading frame interpolation approaches to evaluate our EGVD model's performance. Our comparative analysis includes conventional optical flow-based methods (SuperSloMo~\cite{jiang2018super} and RIFE~\cite{huang2022real}), an event-based approach (CBMNet~\cite{kim2023event}), and a diffusion-based method (DualSVD~\cite{wang2024generative}). To ensure fair comparison, we use officially released pre-trained weights for all baseline methods. For CBMNet, we specifically employ their CBMNet-Large.
%variant trained on the BSRGB dataset. 
Due to DualSVD's resolution sensitivity, we process inputs at 1024×576 resolution before resizing them to 512×512.

\begin{figure*}
    \centering
    \includegraphics[height=0.95\textheight]{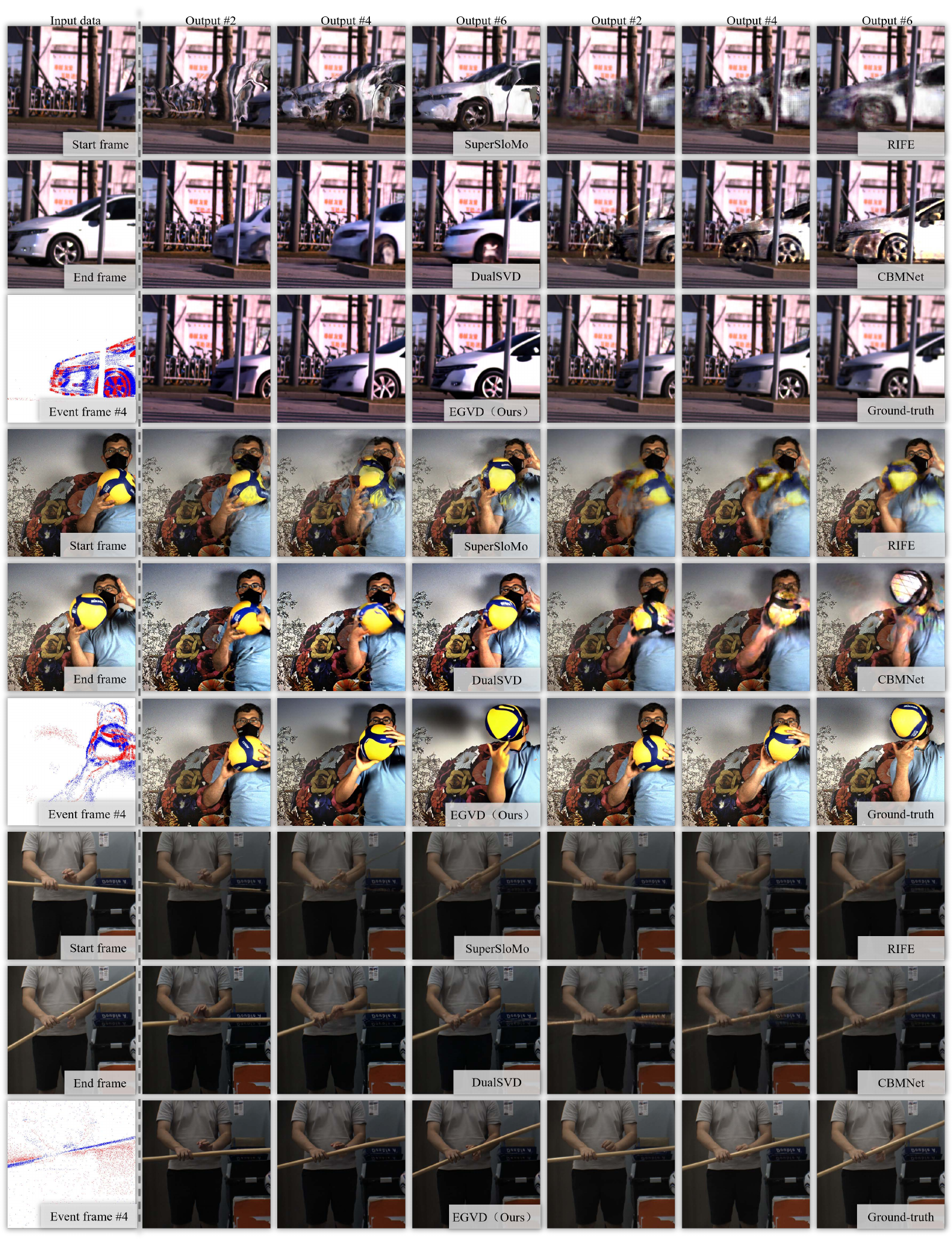}
    \caption{Qualitative comparison of various VFI methods across multiple real scenes. For example, \#i indicates the frame index. The three examples shown from top to bottom are from the real datasets HQEVFI, BSRGB, and ERDS, respectively. Note that all results are generated using a unified set of inference weights, without dataset-specific training. \textbf{See the supplementary video for more results.}}

    \label{fig:real1}
\end{figure*}

\subsubsection{Quantitative results}
\cref{tab:results} presents results across four diverse datasets. On the large-motion DJI 30fps dataset, EGVD demonstrates substantial improvements across all metrics, with remarkable gains in PSNR (+3.73dB) and LPIPS (31.4\% improvement) over the second-best method. This confirms our approach's capability in handling challenging large motion scenarios where conventional techniques fundamentally struggle due to motion ambiguity.

For real-event datasets (Prophesee and BSRGB), our method achieves optimal perceptual quality metrics (LPIPS, MANIQA, MUSIQ, LIQE) while maintaining competitive fidelity measures. Notably, EGVD improves LPIPS by 27.4\% on Prophesee and 24.1\% on BSRGB compared to the next best method, indicating substantially better perceptual quality that aligns with human visual assessment. Even on small-motion GOPRO data where event information provides less distinctive advantage, EGVD outperforms all competitors in PSNR (+1.94dB over RIFE) and SSIM (+0.1129), demonstrating the framework's remarkable versatility across diverse motion scales and scenarios.

\subsubsection{Qualitative analysis}
\cref{fig:real1} illustrates EGVD's significant advantages in challenging scenarios. In the fast-moving vehicle scene (rows 1-3), conventional methods exhibit critical limitations: SuperSloMo generates ghosting artifacts, RIFE produces severe positioning errors, and DualSVD fails to reconstruct correct vehicle geometry. CBMNet better captures position but suffers from detail loss. In contrast, EGVD precisely reconstructs both position and structural details, closely matching ground truth.For non-rigid human motion (rows 4-6), SuperSloMo and RIFE exhibit object deformation and boundary artifacts, DualSVD produces unnatural distortions, and CBMNet introduces significant blur. EGVD reconstructs this challenging motion with exceptional fidelity, preserving both movement trajectory and fine-grained details. In the low-light scene (rows 7-9), conventional methods struggle substantially with ghosting, incorrect motion paths, and blurring. EGVD leverages event data's high dynamic range to overcome these limitations, producing results with accurate positioning, clear edges, and natural appearance even under challenging lighting conditions.

\subsection{Ablation studies}

\begin{table}[t]
\centering
\caption{Ablation study on key components of EGVD framework.}
\label{tab:ablation}
\begin{tabular}{l|ccc}
%\footnotesize
\hline
Configuration & PSNR$\uparrow$ & SSIM$\uparrow$ & LPIPS$\downarrow$ \\
\hline
w/o SVD denoiser & 18.61 & 0.6489 & 0.3859 \\
w/o MMCG & 15.67 & 0.4533 & 0.3425 \\
w/o VFE & 16.95 & 0.5459 & 0.2938 \\
w/o MMF & 20.95 & 0.6788 & 0.1952 \\
w/o RALM & 21.18 & 0.6938 & 0.1857 \\
w/o AW & 18.49 & 0.6879 & 0.2789 \\
\hline
Full model (EGVD) & \textbf{22.25} & \textbf{0.7319} & \textbf{0.1682} \\
\hline
\end{tabular}
\end{table}

\begin{figure}
\setlength{\abovecaptionskip}{0.cm}
\setlength{\belowcaptionskip}{-0.3cm}
    \centering
    \includegraphics[height=0.4\textheight]{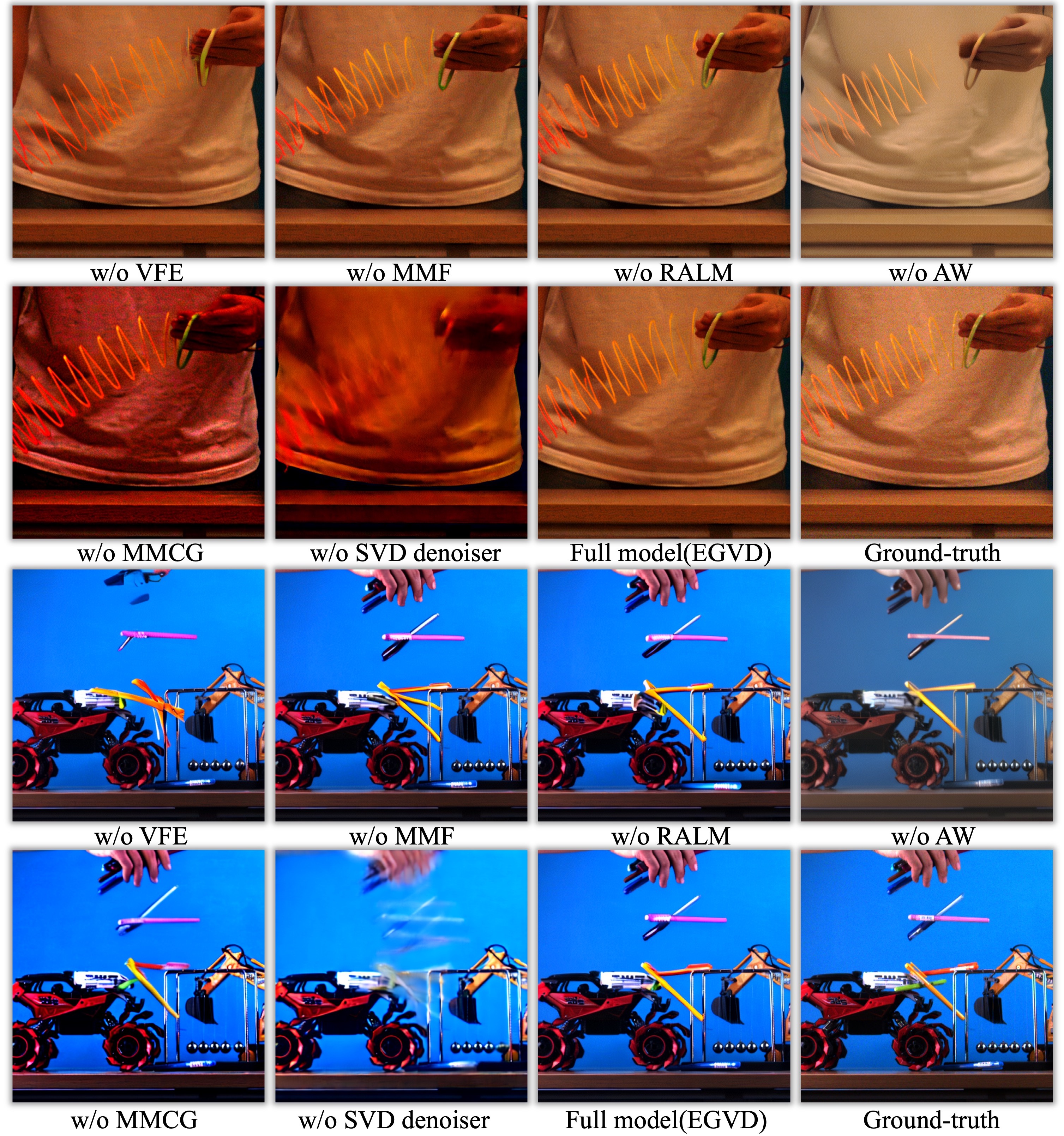}
    \caption{Qualitative ablation study results under different component configurations. (Top: Low-Light, Bottom: Large Motion)}
    \label{fig:ablation}
\end{figure}

To evaluate the contribution of individual components within our EGVD framework, we conduct comprehensive ablation experiments across diverse visual scenes. \cref{tab:ablation} presents quantitative results, while \cref{fig:ablation} provides qualitative visual evidence under challenging conditions.

\textbf{Effect of video diffusion prior:}
To evaluate the impact of the video diffusion prior, we compare our full model with a variant that does not incorporate the video diffusion prior (w/o SVD denoiser). The results demonstrate that our full model achieves significantly better performance on both PSNR and LPIPS. Additionally, in scenes with large motion (\eg, the second image in \cref{fig:ablation}), although events can help 
model (w/o SVD denoiser) recover accurate motion information, the synthesized frames still lack visually pleasing details. This highlights the critical role of diffusion priors in high-quality frame synthesis.

\textbf{Effect of MMCG architecture:}
To validate the effectiveness of the Multi-modal Motion Condition Generator (MMCG), we conduct ablation studies on its individual components. As shown in \cref{tab:ablation} and \cref{fig:ablation}, these studies confirm the effectiveness of our module designs. We focus our discussion on three key components: MMCG as a whole, the Voxel Feature Extractor (VFE), and the Adaptive Weighting (AW) module.

%Removing MMCG entirely—i.e., directly concatenating all available information as the condition to fine-tune SVD—results in a significant performance drop. This is due to the substantial modality gap between the conditioning inputs and the original SVD priors, compounded by the relatively limited data available in Event-VFI. As a result, the model without MMCG (w/o MMCG) suffers from considerable performance degradation. The VFE module plays a crucial role, as its removal leads to a substantial performance drop (PSNR -5.3 dB). Without VFE, the model loses the ability to effectively process and utilize the rich temporal information from event streams, essentially depriving the framework of its core event-based motion cues. Similarly, the AW strategy is essential, as removing it results in a PSNR drop of 3.76 dB. AW fundamentally alters the model’s learning objective—without it, the model must directly predict features in the latent space rather than refining residuals, which significantly increases the learning complexity. This is particularly detrimental in regions with complex motion, leading to less accurate interpolation results.
Removing MMCG entirely, \ie, directly concatenating all inputs to fine-tune SVD, leads to significant performance degradation due to the modality gap and the limited Event-VFI data. The VFE module is crucial for capturing dense temporal motion cues from events, and its absence severely impacts performance. Lastly, AW fundamentally alters the model’s learning objective—without it, the model must directly predict features in the latent space rather than refining residuals, which increases the learning complexity. This is particularly detrimental in regions with complex motion, leading to less accurate interpolation results.

\subsection{Limitations}

%Despite EGVD's visually superior results compared to all existing methods, several limitations remain: our approach yields lower PSNR metrics on certain real-event datasets compared to specialized methods like CBMNet, though perceptual quality is significantly better; computational constraints prevented dataset-specific optimization that could potentially enhance results; and our reliance on simulated events (over half of our training data) introduces a domain gap that could be addressed with more diverse real-world event data. Addressing these limitations in future work could further improve performance across diverse scenarios.

Despite EGVD’s visually superior results over existing methods in large-motion and diverse lighting scenarios, several limitations remain. Due to fidelity issues in the SVD prior, our approach achieves lower PSNR on certain real-event datasets compared to CBMNet, despite offering better perceptual quality. Additionally, the inference speed is limited by the computational cost of SVD.
% \section{Conclusion}

% We presented EGVD, a novel event-guided video diffusion model that effectively addresses frame interpolation in large-motion scenarios by integrating stable video diffusion capabilities with high temporal resolution event data. Our Multi-Modal Motion Condition Generator and selective temporal attention fine-tuning strategy enable the model to generate physically realistic intermediate frames while minimizing training data requirements. Experiments demonstrate significant improvements in perceptual quality across diverse datasets, particularly in challenging scenarios with large motion and varying lighting conditions.

\section{Conclusion}

We presented EGVD, a novel event-guided video diffusion model that effectively addresses frame interpolation challenges in large-motion scenarios by integrating stable video diffusion with high temporal resolution event data. Our Multi-modal Motion Condition Generator fuses RGB and event information to provide precise diffusion guidance, while our selective fine-tuning strategy enables efficient adaptation of pre-trained diffusion priors. By adopting normalization techniques inspired by Elucidated Diffusion Models, we ensure stable training across different noise levels, significantly improving convergence efficiency. Extensive experiments demonstrate that EGVD consistently outperforms existing methods across diverse datasets, particularly in challenging scenarios with large motion and varying lighting conditions. Future work could explore incorporating more diverse real-world event data and extending our approach to other high temporal resolution video tasks.

{
    \small
    \bibliographystyle{ieeenat_fullname}
    \bibliography{main}
}

\end{document}